\documentclass{bmvc2k}

\usepackage{booktabs}
\usepackage{multirow}
\usepackage{color}
\definecolor{mygreen}{RGB}{0,180,0}

\title{MixSIS: Mixed-supervised instance segmentation}

\addauthor{Christian Wilms}{christian.wilms@uni-hamburg.de}{1}
\addauthor{Emre Gülsoylu}{emre.guelsoylu@uni-hamburg.de}{1}
\addauthor{Sven Schreiber}{sven.schreiber@uni-hamburg.de}{1}
\addauthor{Simone Frintrop}{simone.frintrop@uni-hamburg.de}{1}

\addinstitution{
 Computer Vision Group\\
 University of Hamburg\\
 Hamburg, Germany
}

\runninghead{Wilms et al.}{Mixed-supervised instance segmentation}

\begin{document}

\maketitle
\begin{abstract}
Weakly-supervised instance segmentation~(WSIS) based on either box-, point-, or image-level annotations has been proposed to overcome the labor-intensive annotation process in fully-supervised instance segmentation. Despite decent results, most WSIS methods only focus on a single annotation level, ignoring that some objects are easy to spot from point annotations, while others require box or even mask annotations. Therefore, we propose the \textbf{Mix}ed-\textbf{S}upervised \textbf{I}nstance \textbf{S}egmentation~(MixSIS) task, which allows a flexible mix of annotation levels~(image, point, box, and mask) within a dataset. MixSIS has a high practical relevance, as it allows optimizing the annotation process by annotating challenging objects in more detail, while saving time on simpler instances with less detailed annotations. To address MixSIS, we propose  AnnotationsMixer~(AnnsMixer) as a baseline. AnnsMixer is  a truly mixed-supervised instance segmentation system, which generates pseudo mask annotations from all annotation levels, even mixed within a single image. We evaluate AnnsMixer on both WSIS and MixSIS tasks, largely outperforming state-of-the-art methods. In three case studies, we also demonstrate how the unique features of MixSIS and AnnsMixer can substantially reduce the annotation time~(-90\%) or improve the results~(+34\%) in instance segmentation with a simple annotation rule or leveraging heterogeneous data. Our code is available at \url{https://github.com/chwilms/MixSIS}.
\end{abstract}


\section{Introduction}
\begin{figure}
    \centering
    \includegraphics[width=.8\linewidth]{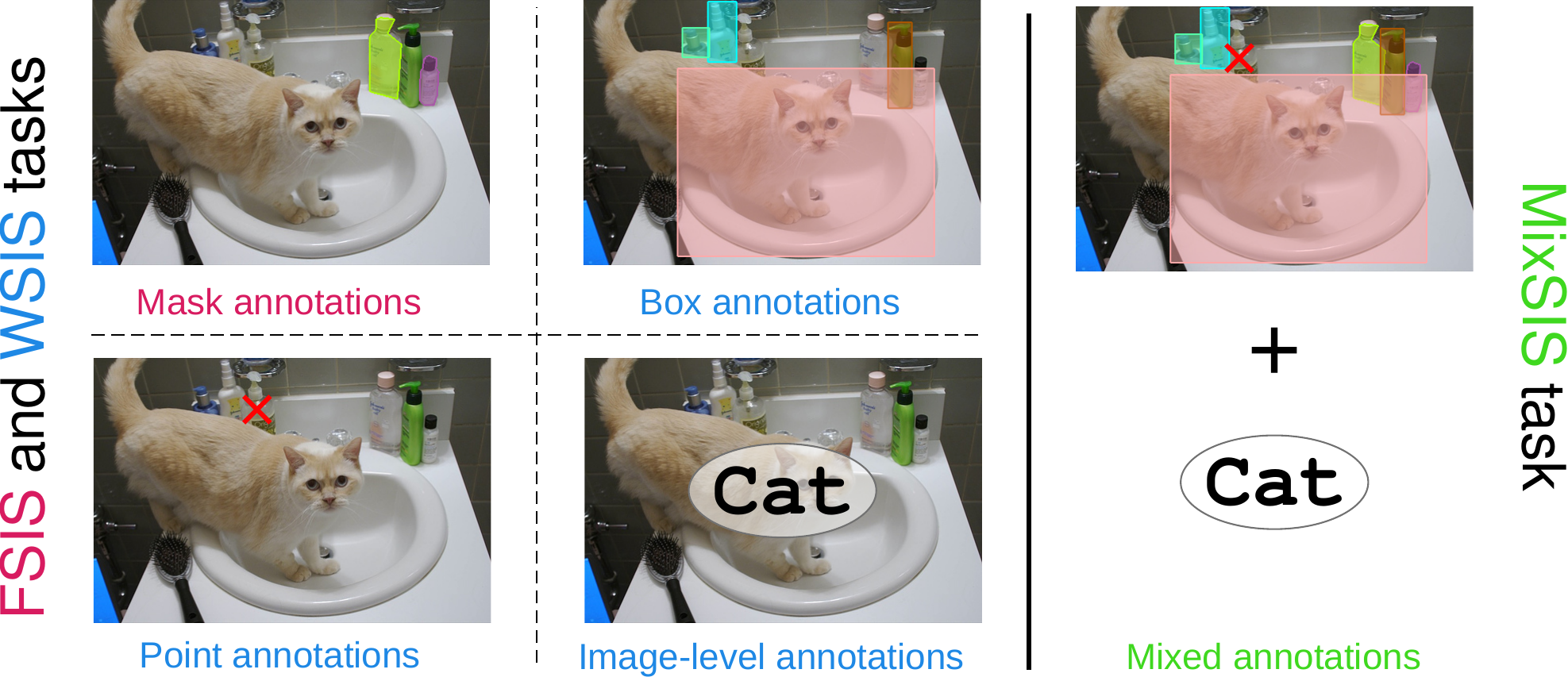}
    \caption{Comparison of the FSIS and WSIS tasks~(left), where all instances are labeled on the same annotation level, and the MixSIS task~(right), where all annotation levels are combined even within the same image.}
    \label{fig:coverFigTasks}
\end{figure}

Fully-supervised instance segmentation~(FSIS) based on manually annotated segmentation masks is costly, due to the labor-intensive annotation process~\cite{kim2023devil}. To mitigate this practically important issue, weakly-supervised instance segmentation~(WSIS) was proposed~\cite{cheng2023boxteacher,wei2024semantic,lan2021discobox,kim2022beyond}. In WSIS, depending on the subtask, all instances are labeled on the same annotation level using either box, point, or image-level annotations as depicted in Fig.~\ref{fig:coverFigTasks}.


\begin{figure}
    \centering
    \begin{tabular}{cc}
    \includegraphics[width=0.35\linewidth]{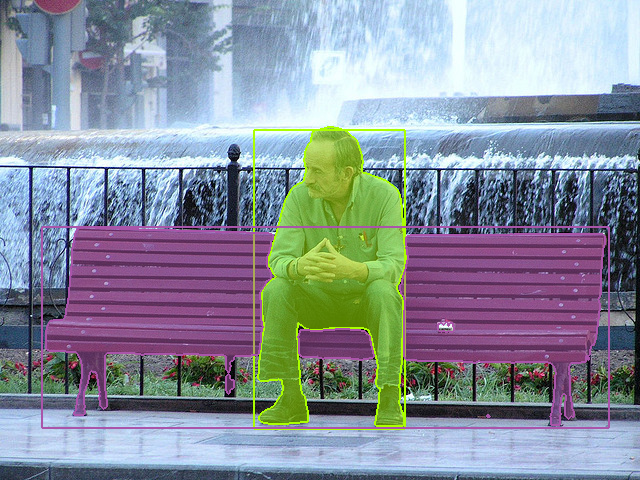} & 
    \includegraphics[width=0.35\linewidth]{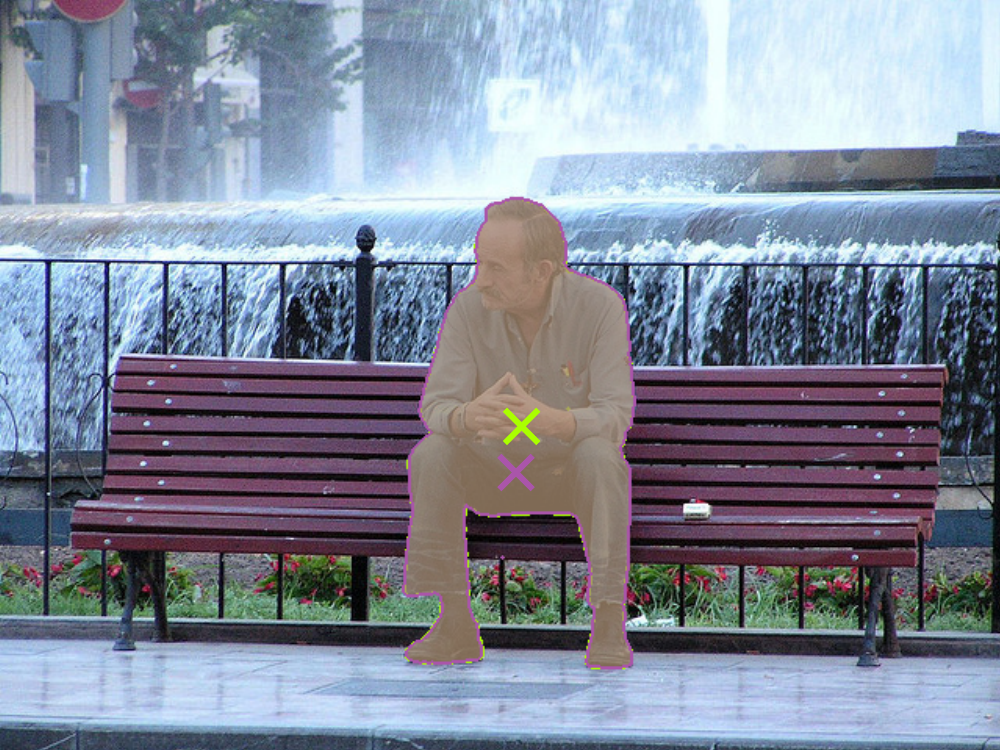}
    \end{tabular}
    \caption{Pseudo mask annotations from box~(left) or point~(right) annotations generated with our AnnsMixer. While the mask for the person is accurate from a box or a point, the bench needs a box, as the point annotation falls within the person, making it ambiguous.}
    \label{fig:coverFigAnns}
\end{figure}

As objects are diverse within a dataset and even an image, some objects are easily captured with less supervision, like a point annotation, while others significantly profit from stronger supervision, like a box or mask. For instance, annotating the bench in Fig.~\ref{fig:coverFigAnns} with its centroids fails as the person's centroid is very similar. Using a box in this case solves the issue. Note that the person here is properly captured from its centroid only. Hence, investing additional annotation time for the bench is beneficial, while the person only needs a point annotation. Using a point instead of a box reduces the annotation time by 65\%~\cite{kim2023devil}. Overall, selecting a single proper supervision level for a dataset at hand is challenging due to the diverse object or scene properties. This leads to suboptimal segmentation results or excessive annotation efforts.

To circumvent this problem, only a few studies have combined FSIS and WSIS~\cite{bellver2019budget,kim2023devil,kweon2025wish}. These studies either fuse full supervision with a single level of weak supervision, e.g., masks and points~\cite{kim2023devil}, or restrict to a single level of weak supervision for all instances of an image. However, this does not allow tailoring the annotation level to each object according to its specific properties. Hence, a flexible alternative is needed, which allows combining all common annotation-levels even within a single image. This offers the advantage of annotating objects based on their individual properties and not limiting the annotation process to one or two annotation levels per dataset or a single annotation level per image. Hence, segmentation results and the annotation process are both optimized.

In this paper, we fill this void by first defining the new \textbf{Mix}ed-\textbf{S}upervised \textbf{I}nstance \textbf{S}egmentation~(MixSIS) task, which is a flexible mixture of fully-supervised and all levels of weakly-supervised instance segmentation~(see Fig.~\ref{fig:coverFigTasks}). Thus, MixSIS allows combining all supervision-levels within a single image. To address MixSIS, we propose the pioneering system AnnotationsMixer~(AnnsMixer) as a baseline. AnnsMixer produces pseudo mask annotations given box-, point-, or image-level annotations to train a fully-supervised instance segmentation system. We compare AnnsMixer with state-of-the-art WSIS systems on both the WSIS and the MixSIS task, showcasing its strong performance, largely outperforming state-of-the-art. To demonstrate the practical relevance of MixSIS's and AnnsMixer's unique features, we present multiple case studies. First, we design an efficient annotation scheme for instance segmentation, reducing the annotation time by up to 90\% with a simple annotation rule, while only suffering a minor degradation in performance. Moreover, we exploit additional data sources for instance segmentation in multiple ways using MixSIS and AnnsMixer to improve the results by up to 34\%. 

In summary, our contributions are threefold: 
\begin{enumerate}
    \item Introducing MixSIS task with pioneering baseline system AnnsMixer to produce state-of-the-art results on WSIS and MixSIS.
    \item Leveraging the MixSIS framework and AnnsMixer to propose an enhanced annotation scheme, reducing the annotation time by up to 90\%.
    \item Applying AnnsMixer in a MixSIS framework to improve instance segmentation results by up to 34\%, highlighting its practical relevance.
\end{enumerate}

\section{Related Work}

\subsection{Weakly-supervised Instance Segmentation}

Weakly-supervised instance segmentation (WSIS) with box-~\cite{tian2021boxinst,lan2021discobox,li2022box,cheng2023boxteacher,li2023sim}, point-~\cite{laradji2020proposal,liao2023attentionshift}, or image-level~\cite{ahn2019weakly,arun2020weakly,kim2022beyond} supervision has been extensively studied. For box-level supervision, several approaches use multiple instance learning and specialized loss functions in fully-supervised instance segmentation systems~\cite{tian2021boxinst,lan2021discobox}. 
Other approaches include level-sets~\cite{li2022box} or prototype features~\cite{li2023sim}. Employing image-level supervision, most systems rely on activation maps of image classifiers to locate objects. 
In \cite{ahn2019weakly},~peaks of class activation maps (CAMs)~\cite{zhou2016learning} are used to locate objects and infer the segmentation from pixel affinity. Recently, \cite{zhu2024weaksam}~applied SAM to generate pseudo annotations in a multi-stage pipeline utilizing an additional object detector. 
Besides systems based on box- and image-level supervision, few systems address single-point supervision. \cite{laradji2020proposal}~learn to locate a single object point and combine object embeddings with traditional object proposals, while \cite{liao2023attentionshift}~use self-attention maps in vision transformers to iteratively capture object parts.   

Overall, WSIS approaches typically use a single level of weak annotations and employ complex training schemes. In contrast, our AnnsMixer uses a simple and unified approach and works with arbitrary instance segmentation systems to handle all levels of weak and strong annotations jointly.

\subsection{Mixed-supervised Instance Segmentation}

Few WSIS approaches allow multiple annotation levels~\cite{bellver2019budget,kim2023devil,kweon2025wish}. \cite{bellver2019budget}~combine mask- and image-level annotations by training an LSTM-based pseudo annotation generator, conditioned on the image-level annotations. Combining point and mask annotations, \cite{kim2023devil}~propose a teacher-student training scheme, where the teacher learns to generate segmentation masks given both annotations as ground truth. The teacher then creates pseudo annotations for the student network. Recently, \cite{kweon2025wish}~leverage all levels of weak supervision based on a hardcoded combination of Mask2Former and SAM. While allowing all annotation levels, they are not mixed within an image, which does not allow choosing the most efficient annotation type per instance, but only per image. Moreover, \cite{kweon2025wish}~does not support mask annotations and is bound to Mask2Former through a complex multi-stage matching. Overall, little work has been conducted to address WSIS in a fully flexible manner, mixing all annotation levels freely within an image.




\section{Mixed-supervised Instance Segmentation}
\label{sec:task}

To overcome the reliance on a single annotation level, we propose the new \textbf{Mix}ed-\textbf{S}upervised \textbf{I}nstance \textbf{S}egmentation~(MixSIS) task. MixSIS is an instance segmentation task, where each instance is either annotated with a mask, a box, a point, or on image-level. Further, the annotation levels can be freely mixed within each image and across categories of a dataset. This flexible mix of annotation levels is the key difference between MixSIS and WSIS and allows various applications like efficient annotation time management or mixing heterogeneous annotations~(see Sec.~\ref{sec:apps}). However, it demands systems to handle all annotation levels to fully leverage the annotations.

To evaluate MixSIS, we use datasets for fully-supervised instance segmentation like COCO~\cite{lin2014microsoft}. Hence, the amount of annotations per level is fully flexible, since we can derive each annotation level from a mask~(e.g., we use the centroid of the mask for point supervision). To assign each annotated instance to an annotation level, i.e., mask, box, point, or image, we propose random mixes of different complexities. In particular, we define an even mix, where 25\% of the annotations are assigned to each level. We coin this Mix 1 (25/25/25/25), where each number denotes the percentage of mask-, box-, point-, and image-level annotations. Mix 2 (10/20/30/40) and Mix 3 (5/10/25/60) are more challenging with a stronger focus on weak supervision and reduced annotation effort. For testing, we utilize mask annotations as in WSIS.

Besides these random mixes, designed mixes that reduce the annotation time while maintaining strong detection results are possible based on the different characteristics of the annotated objects. For instance, compact objects may only need point annotations, while less compact objects require more supervision. This is one of the major advantages of MixSIS over WSIS. We will discuss this in Sec.~\ref{sec:effAnns} and propose an efficient annotation mix.

\section{AnnsMixer for MixSIS}
\begin{figure}[!t]
    \centering
    \includegraphics[width=0.98\linewidth]{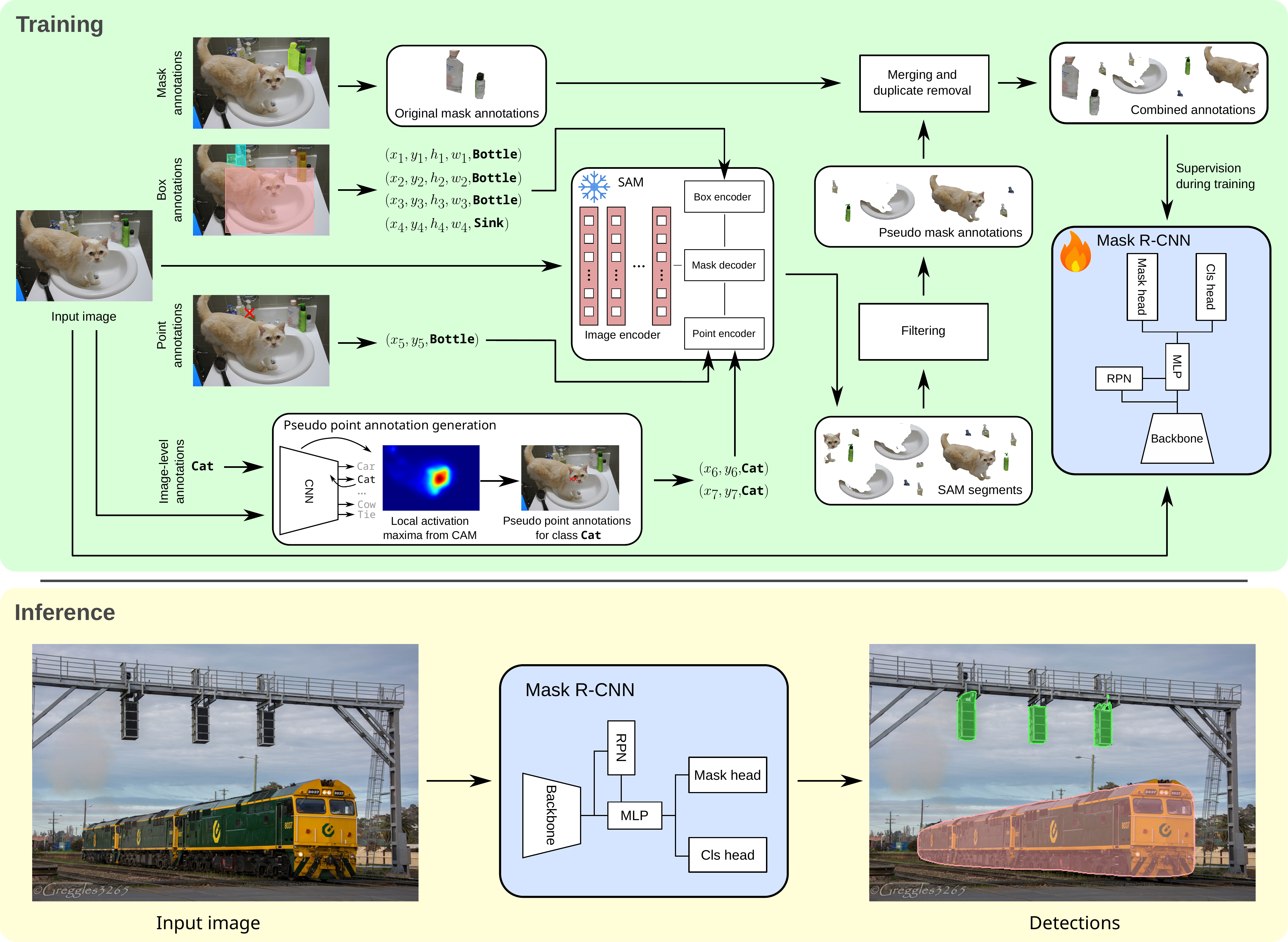}
    \caption{Overview of our pioneering baseline system AnnsMixer for MixSIS and WSIS. Prompts are derived from box-, point-, and image-level annotations to generate pseudo mask annotations using SAM. After combining them with original mask annotations, a vanilla Mask R-CNN is trained and, subsequently, applied during inference.} 
    
    \label{fig:sysFig}
\end{figure}

We propose the pioneering system AnnsMixer~(see Fig.~\ref{fig:sysFig}) as a baseline to address MixSIS. The core idea of AnnsMixer is to generate pseudo mask annotations from any annotation level. To generate pseudo mask annotations, AnnsMixer applies SAM~\cite{kirillov2023segment} with geometric prompts derived from weak annotations. Subsequently, these pseudo mask annotations are combined with original mask annotations, if available, and used to supervise a vanilla fully-supervised instance segmentation system like Mask R-CNN~\cite{he2017mask}. During inference~(yellow box in Fig.~\ref{fig:sysFig}), this instance segmentation system is applied.

\subsection{Generating Pseudo Mask Annotations}
AnnsMixer starts by discerning the annotations by their level~(see horizontal streams in Fig.~\ref{fig:sysFig}). Mask annotations are directly added to the final set of annotations. For weak annotations, geometric prompts, i.e., boxes or points, are necessary to prompt SAM. For box and point annotations~(second and third streams in Fig.~\ref{fig:sysFig}), the annotations themselves are the prompts. Image-level annotations~(fourth stream in Fig.~\ref{fig:sysFig}) have to be transformed into geometric prompts.  We follow~\cite{ahn2019weakly} and ground the annotated object classes by analyzing CAMs~\cite{zhou2016learning} of an image classifier trained on the respective dataset. Hence, the classifier has been trained in a multi-class fashion, only detecting the existence of an object class in an image without any explicit localization. 
From the CAM of an annotated class, we extract the coordinates of $n$ local maxima as object locations and point prompts. Note that we set $n=2$ in our experiments. This process is visualized in the bottom left box in the green part of Fig.~\ref{fig:sysFig}. Overall, each weak annotation is now represented by one or multiple geometric prompts.

Subsequently, we prompt a pre-trained SAM with these geometric prompts to produce segments. Using SAM here exploits the rich knowledge encoded in the foundation model. We allow SAM to generate three segments per prompt to account for ambiguous prompts~\cite{kirillov2023segment}. This yields a set of class-agnostic SAM segments, as visualized in the center of Fig.~\ref{fig:sysFig}. Class-agnostic refers to the fact that SAM does not classify the segments; however, the segments still carry their originally annotated object class label. Since the segments may be incomplete or noisy, we remove low-confidence segments in a filtering step through thresholding based on the segments' stability score and estimated Intersection over Union (IoU) values. Both values are outputs from SAM. The remaining segments are the pseudo mask annotations, which are added to the original mask annotations, before non-maximum suppression removes near duplicates. The result is a set of combined mask annotations.


\subsection{Instance Segmentation Process}
Given these annotations, we train Mask R-CNN~\cite{he2017mask} in a 'fully'-supervised manner, which means that all annotations are now represented by masks, while originally some of them were weak annotations. Note that any fully-supervised instance segmentation system is possible here, and we leave it as future work to integrate more recent models as our system acts as a proof of concept. Generally, this modularity offers great flexibility and enables us to integrate future instance segmentation models into our baseline system, AnnsMixer. For training Mask R-CNN, we follow the default settings~\cite{he2017mask}. Note that during inference, only the trained Mask R-CNN is applied~(see the yellow part in Fig.~\ref{fig:sysFig}).

\section{Evaluation}

To assess the quality of our baseline system AnnsMixer and understand the performance of WSIS systems on our novel MixSIS task, we evaluate on the COCO dataset~\cite{lin2014microsoft}. We follow our MixSIS evaluation protocol~(see Sec.~\ref{sec:task}). Thus, we train on a variety of annotation levels within one experiment, given by the random mixes. We compare AnnsMixer with WSIS systems, which only utilize a part of the annotations, as they are restricted to one type of annotations. For improved comparability, we transform more detailed annotations~(e.g., boxes) into less detailed ones~(e.g., points). To overcome this structural disadvantage for WSIS systems, we also present the results of AnnsMixer in the traditional WSIS setting using either box-, point-, or image-level annotations. Here, all annotations are given on the same annotation level. For both tasks, we train on the COCO \textit{train} split and evaluate on the COCO \textit{val}~(MixSIS) or \textit{test-dev}~(WSIS) splits. The results are reported in terms of mean Average Precision~(mAP).

\subsection{Mixed-supervised Instance Segmentation}
\label{sec:resMixSIS}

For the new MixSIS task, we compare AnnsMixer to Mask R-CNN~\cite{he2017mask}, utilizing only mask annotations, BoxLevelSet~\cite{li2022box} and MAL~\cite{lan2023vision}, which utilize mask and box annotations as boxes, and SAPNet~\cite{wei2024semantic} leveraging mask, box, and point annotations as points We selected these methods as they provide top results in the WSIS setup~(see Tab.~\ref{tab:resCOCO}). Note that all results of image-level WSIS systems~(see Tab.~\ref{tab:resCOCO}) are directly comparable with MixSIS results as all annotations will fall back to the image-level for these systems. Comparing against~\cite{kweon2025wish}, who handle multiple annotation levels, is impossible as no code is available. Table~\ref{tab_mixSIS} presents the MixSIS results of all systems on the COCO \textit{val} split. AnnsMixer shows strong MixSIS performance and outperforms all other methods in mAP. We attribute this to the larger amount of usable annotations compared to all other models, which only focus on one annotation level. For instance, fully-supervised Mask R-CNN can only leverage 25\%~(Mix 1) to 5\%~(Mix 3) of the annotations, which degrades the performance by up to 0.098 in mAP. The box-based systems MAL and BoxTeacher can utilize up to 50\% of the annotations, leading to substantially improved results over fully-supervised Mask R-CNN. Yet, our baseline system AnnsMixer still outperforms. Finally, SAPNet utilizes all annotations as points except image-level annotations. With its specialized architecture tailored to point annotations, it is closest to flexible AnnsMixer but still performs worse in mAP.

\begin{table}[h]
\centering
\caption{MixSIS results on the COCO \textit{val} split. The tuple ($\mathcal{M}$/$\mathcal{B}$/$\mathcal{P}$/$\mathcal{I}$) indicates the percentage of mask-, box-, point-, and image-level annotations. Best results  are in bold. $^\dagger$ denotes annotations transformed to boxes or points to match the systems' inputs.}
\label{tab_mixSIS}
{
    \resizebox{\linewidth}{!}{
    \begin{tabular}{@{}llccccc@{}}
    \toprule
     & \textbf{System} & Mask R-CNN \cite{he2017mask} & MAL~\cite{lan2023vision} & SAPNet~\cite{wei2024semantic} & BoxLevelSet \cite{li2022box} & AnnsMixer (our) \\
     & \textbf{Backbone} & R-50 & R-50 & R-50 & R-101 & R-50 \\
    \textbf{Setup} & \textbf{Supervision} & $\mathcal{M}$ & $\mathcal{M^\dagger,B}$ & $\mathcal{M^\dagger,B^\dagger,P}$ & $\mathcal{M^\dagger,B}$ & $\mathcal{M,B,P,I}$ \\ \midrule
    \multirow{3}{*}{\begin{tabular}[c]{@{}l@{}}Mix 1\\ (25/25/25/25)\end{tabular}} & mAP & 0.239 & 0.317 & 0.297 & 0.277 & \textbf{0.323} \\
     & mAP$_{50}$ & 0.405 & 0.517 & 0.500 & 0.489 & \textbf{0.523} \\
     & mAP$_{75}$ & 0.249 & 0.335 & 0.304 & 0.277 & \textbf{0.344} \\ \midrule
    \multirow{3}{*}{\begin{tabular}[c]{@{}l@{}}Mix 2\\ (10/20/30/40)\end{tabular}} & mAP & 0.196 & 0.288 & 0.285 & 0.234 & \textbf{0.295} \\
     & mAP$_{50}$ & 0.334 & 0.473 & \textbf{0.485} & 0.426 & 0.480 \\
     & mAP$_{75}$ & 0.205 & 0.303 & 0.291 & 0.228 & \textbf{0.315} \\ \midrule
    \multirow{3}{*}{\begin{tabular}[c]{@{}l@{}}Mix 3\\ (5/10/25/60)\end{tabular}} & mAP & 0.168 & 0.245 & 0.263 & 0.172 & \textbf{0.266} \\
     & mAP$_{50}$ & 0.290 & 0.407 & \textbf{0.452} & 0.317 & 0.435 \\
     & mAP$_{75}$ & 0.173 & 0.258 & 0.268 & 0.166 & \textbf{0.284} \\ \bottomrule
    \end{tabular}
    }
}
\end{table}

In addition to this evaluation based on the three proposed mixes, we also investigate the influence of each annotation level more closely. For this analysis, we start from 25\% mask annotations only and add either 75\% point, box, or image-label annotations. Note that the initial setup with 25\% mask annotations only resembles the Mask R-CNN result from Tab.~\ref{tab_mixSIS}. The results of AnnsMixer with the additional annotations in Tab.~\ref{tab:mixSISabl} clearly indicate that all levels of weak annotations improve over the mask-only baseline. Hence, the formulation of the MixSIS task, combining any of these annotation levels, is reasonable. Unsurprisingly, box-annotations lead to the best results, with point annotation ending up second. Interestingly, the results of AnnsMixer in the Mix 1 setup with 25\% annotations from each level are just short of adding the costly 75\% box annotations to the 25\% mask annotations.

\begin{table}[h]
\centering
\caption{MisSIS results on COCO \textit{val} split with different combinations of annotations levels. The first and last row are the respective systems' setups from Mix 1 in Tab.~\ref{tab_mixSIS}.}
\label{tab:mixSISabl}
\begin{tabular}{@{}lcccc@{}}
\toprule
System & Supervision         & mAP & mAP$_{50}$ & mAP$_{75}$ \\ \midrule
Mask R-CNN & 25\%$\mathcal{M}$       & 0.239 & 0.405 & 0.249 \\
\midrule
AnnsMixer (our) & 25\%$\mathcal{M}$+75\%$\mathcal{I}$& 0.267 & 0.429 & 0.288 \\
AnnsMixer (our) & 25\%$\mathcal{M}$+75\%$\mathcal{P}$& 0.298 & 0.494 & 0.313 \\
AnnsMixer (our) & 25\%$\mathcal{M}$+75\%$\mathcal{B}$& 0.346 & 0.561 & 0.367 \\
\midrule
AnnsMixer (our) & 25\%$\mathcal{M}$+25\%$\mathcal{B}$+25\%$\mathcal{P}$+25\%$\mathcal{I}$& 0.323 & 0.523 & 0.344 \\
\bottomrule
\end{tabular}
\end{table}

\subsection{Weakly-supervised Instance Segmentation}

We also evaluate AnnsMixer in the traditional WSIS setups with either box-, point-, or image-level annotations, comparing it to several recent WSIS systems. Note that here, all systems are able to utilize all annotations, but our system is the only one not tailored to a specific annotation level. The results in Tab.~\ref{tab:resCOCO} on the COCO \textit{test-dev} split indicate the strong performance of our AnnsMixer compared to dedicated WSIS systems. While being second in the point- and box-supervised cases, it even outperforms dedicated WSIS methods for image-level supervision. Compared to other WSIS systems using SAM, WISH~\cite{kweon2025wish} and WeakSAM~\cite{zhu2024weaksam}, AnnsMixer performs better given the same backbone. Overall, AnnsMixer, developed as a baseline system for MixSIS, shows very competitive results on the WSIS task without an architecture dedicated  to a single annotation level. 

\begin{table}[tbp]
\centering
\caption{WSIS results on the COCO \textit{test-dev} split. $\mathcal{M},\mathcal{B},\mathcal{P},\text{ and }\mathcal{I}$ denote mask-, box-, point-, and image-level supervision.  Best results per level are in bold. $^\dagger$: Results on the COCO \textit{val} split.}
\label{tab:resCOCO}
{\small
\begin{tabular}{@{}lccccc@{}}
\toprule
System & Backbone & Supervision & mAP & mAP$_{50}$ & mAP$_{75}$   \\ \midrule
Mask R-CNN \cite{he2017mask}  & ResNet-50 & $\mathcal{M}$ & 35.5 & 57.0 & 37.8 \\ 
\midrule
\midrule
BoxInst \cite{tian2021boxinst} & ResNet-50 & $\mathcal{B}$ & 32.1 & 55.1 & 32.4 \\ 
DiscoBox \cite{lan2021discobox} & ResNet-50 & $\mathcal{B}$ & 32.0 & 53.6 & 32.6 \\ 
BoxTeacher \cite{cheng2023boxteacher} & ResNet-50 & $\mathcal{B}$ & 32.9 & 54.1 & 34.2 \\ 
MAL~\cite{lan2023vision} & ResNet-50 & $\mathcal{B}$ & \textbf{35.7} & - & -  \\ 
BoxSeg~\cite{lai2025boxseg} & ResNet-101 & $\mathcal{B}$ & 33.3 & 53.4 & 35.0  \\ 
BoxInst \cite{tian2021boxinst} & ResNet-101 & $\mathcal{B}$ & 32.5 & 55.3 & 33.0 \\ 
BoxLevelSet \cite{li2022box} & ResNet-101 & $\mathcal{B}$ & 33.4 & \textbf{56.8} & 34.1 \\ 
SIM \cite{li2023sim} & ResNet-101 & $\mathcal{B}$ & 34.0 & \textbf{56.8} & 35.0  \\ 
\midrule
AnnsMixer (our)  & ResNet-50 & $\mathcal{B}$ & 34.5 & 56.2 & \textbf{36.5}  \\ 
\midrule
\midrule
BESTIE~\cite{kim2022beyond} & HRNet48 & $\mathcal{P}$ & 17.8 & 34.1 & 16.7 \\ 
AttnShift~\cite{liao2023attentionshift} & ViT & $\mathcal{P}$ & 21.9 & 43.5 & 20.1 \\
P2MNet-MR~\cite{chen2025p2object} & ResNet-50 & $\mathcal{P}$ & 25.6 & 48.4 & - \\
SAPNet~\cite{wei2024semantic} & ResNet-50 & $\mathcal{P}$ & \textbf{31.2}$^\dagger$ & \textbf{51.8}$^\dagger$ & \textbf{32.3}$^\dagger$ \\
\midrule
AnnsMixer  (our)  & ResNet-50 & $\mathcal{P}$ & 26.2 & 44.9& 27.2 \\ 
\midrule
\midrule
WISH~\cite{kweon2025wish} &ResNet-50 & $\mathcal{I}$ & 20.9 & - & - \\
WeakSAM~\cite{zhu2024weaksam} &ResNet-50 & $\mathcal{I}$ & 21.0 & 34.5 & 22.2 \\
\midrule
AnnsMixer  (our) & ResNet-50 & $\mathcal{I}$ & \textbf{22.0} & \textbf{35.6} & \textbf{23.6}  \\ 
 \bottomrule
\end{tabular}
}
\end{table}

\subsection{Quality of Generated Pseudo Mask Annotations}
\label{sec:eval_gt}
To further show the quality of AnnsMixer's pseudo mask annotations, we compare our pseudo mask annotations with the original mask annotations on the COCO training dataset. Table~\ref{tab:resGT} presents the results in terms of IoU, recall, and precision between original mask annotations and pseudo mask annotations generated in AnnsMixer.  Our box-based pseudo mask annotations match the original annotations very well. Across the dataset, $84.1\%$ of the originally annotated masks are matched by the pseudo mask annotations, while $95.7\%$ of our annotations cover an original annotation. Overall, the IoU in the case of box-based pseudo mask annotations is 70.5, highlighting the high quality of our generated annotations again. For point- and image-level supervision, the generated masks degrade due to missing spatial extent~(point- and image-level) and localization~(image-level). Still,  $66.8\%$ of the point-based masks and $48.5\%$ of the masks based on image-level annotations match the original annotations. Despite the drop, some objects are still suitable for these annotation levels as we will discuss in Sec.~\ref{sec:effAnns}.

\begin{table}
\centering
\caption{Comparison of original ground truth masks and generated pseudo mask annotations in our proposed AnnMixer given weak supervision levels~($\mathcal{B},\mathcal{P}, \text{ and } \mathcal{I}$) on COCO \textit{train} split. Recall and precision are calculated based on an IoU of 50~\cite{everingham2010pascal}.}
\label{tab:resGT}
\begin{tabular}{@{}cccc@{}}
\toprule
Supervision & IoU & Recall & Precision \\ \midrule 
$\mathcal{B}$ & 70.5 & 84.1\% & 95.7\% \\ 
$\mathcal{P}$ & 56.6 & 62.8\% & 66.8\% \\ 
$\mathcal{I}$ & 27.1 & 29.3\% & 48.5\% \\  \bottomrule
\end{tabular}
\end{table}

\begin{figure}[tb]
\centering
\begin{tabular}{cccc}
\includegraphics[width = .21\linewidth]{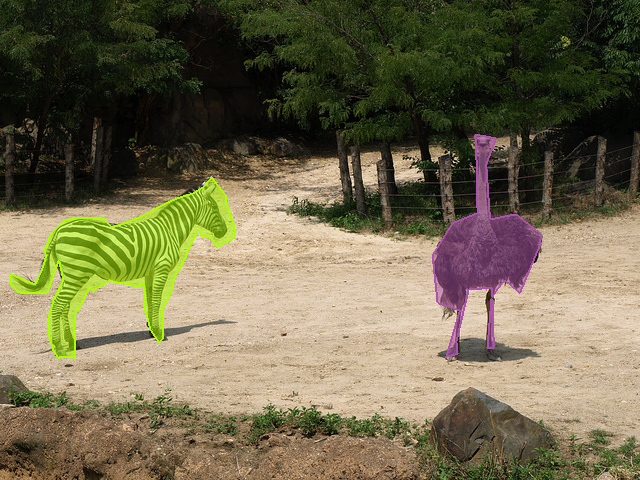} &
\includegraphics[width = .21\linewidth]{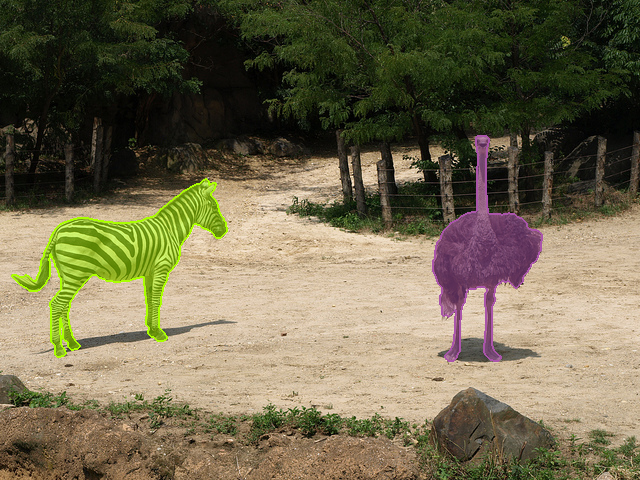} &
\includegraphics[width = .21\linewidth]{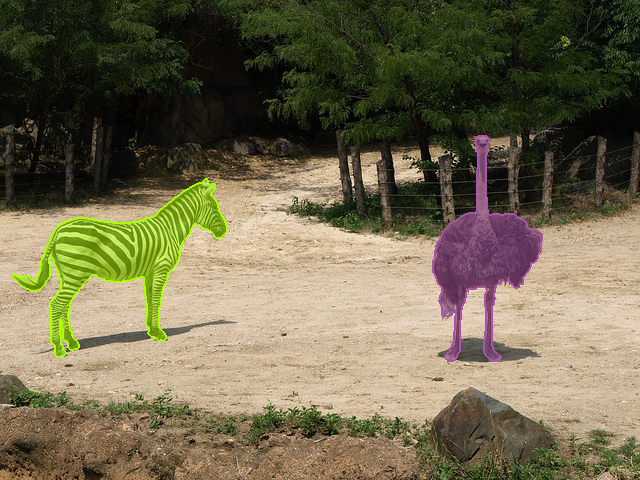} &
\includegraphics[width = .21\linewidth]{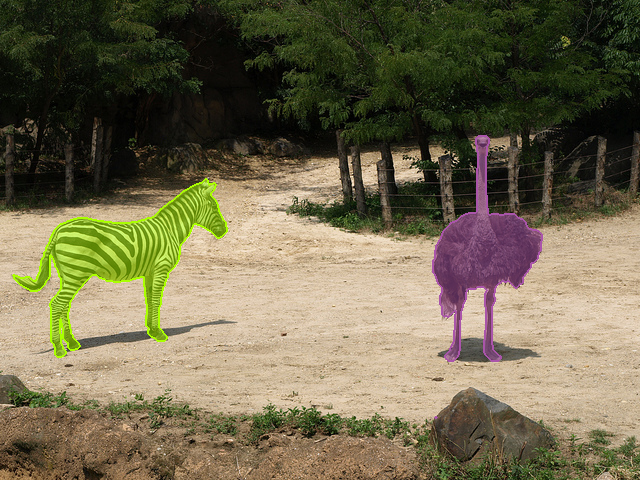}\\
\includegraphics[width = .21\linewidth]{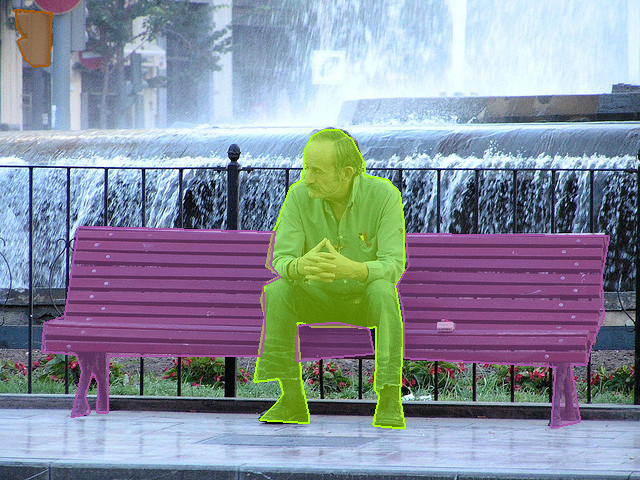} &
\includegraphics[width = .21\linewidth]{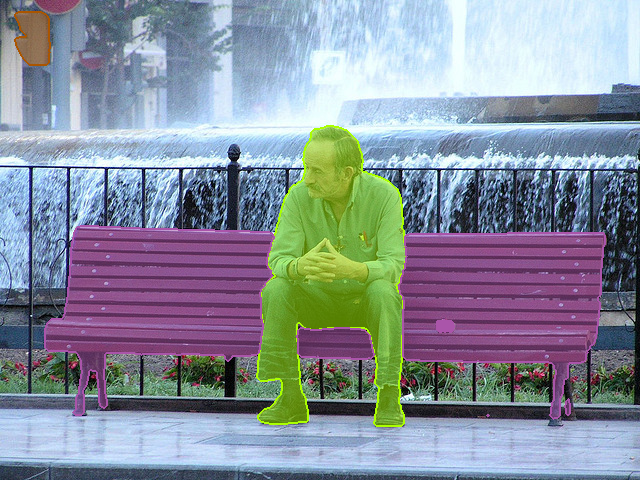} &
\includegraphics[width = .21\linewidth]{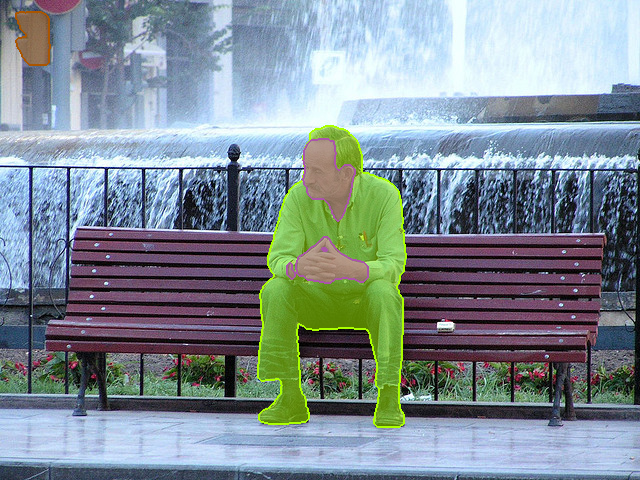} &
\includegraphics[width = .21\linewidth]{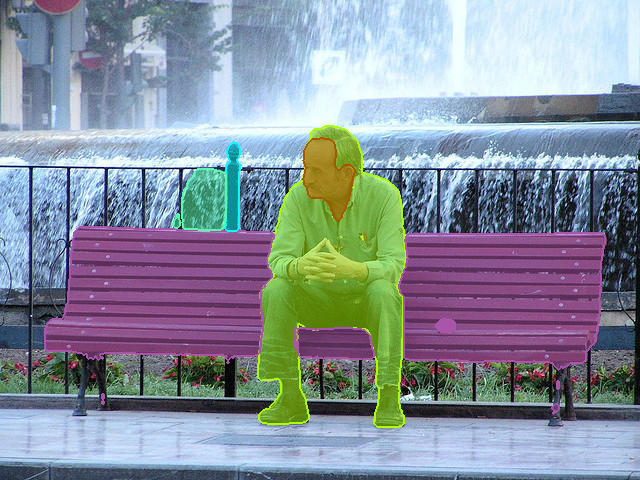}\\

{\small COCO masks} &{\small Masks from}&{\small Masks from}&{\small Masks from}\\
{\small } &{\small box-level}&{\small point-level}&{\small image-level}
\end{tabular}
\caption{Examples of original COCO mask annotations and pseudo mask annotations from box-, point-, or image-level supervision generated in our proposed AnnsMixer. We use the centroid of the COCO mask annotation as point supervision.}
\label{fig:qualGT}
\end{figure}

Figure~\ref{fig:qualGT} complements these results, comparing original COCO annotation masks with the masks generated in AnnsMixer. In the first row, it is visible that all supervision levels lead to high-quality pseudo mask annotations in AnnsMixer, even surpassing the original masks' quality in details along the objects' contours. The lower row shows typical problems related to point- and image-level supervision. First, the centroid of the bench is located within the person; hence, the point-level masks miss the bench. Moreover, both point- and image-level supervision generate masks for object parts like the person's face. For image-level supervision, this is due to the strong influence of faces on the classification result for class \texttt{Person}. This leads to a high class activation in the face area. For point-level supervision, both hands and face are detected as one segment due to their visual similarity. Overall, this analysis shows the high quality of the generated annotations in AnnsMixer, especially for box-level supervision.

\section{Case Studies}
\label{sec:apps}

We investigate the practical application of mixing annotation levels in instance segmentation in three case studies, which highlight the strength of our MixSIS formulation and the pioneering baseline system AnnsMixer.

\subsection{Efficient Annotation with MixSIS}
\label{sec:effAnns}
In the first case study, we investigate efficient annotation schemes based on the MixSIS task formulation. Instead of randomly assigning instances to an annotation level as in Sec.~\ref{sec:resMixSIS} or allowing only a single annotation level, we design a simple annotation scheme that takes the properties of objects into account to decide which annotation level is sufficient for each object. The idea is to optimize the tradeoff between annotation time and detection performance.

\subsubsection{Object Properties}
For developing an efficient annotation scheme for the MixSIS task, we evaluated several object properties as possible indicators for assigning objects to mask-, box-, point-, or image-level supervision. To automate the process of estimating the object properties during development of the scheme, we selected the COCO \textit{train} split with its pixel-precise mask annotations as a basis. From these masks, we automatically derived all object properties. Note that in real applications, human annotators will easily decide on the objects' properties, which are natural to humans~(see Fig.~\ref{fig:visGoodCont}). The object properties are grouped into four categories. 

The properties in the first category are \textit{boxness}, \textit{center distance}, \textit{eccentricity}, \textit{circularity}, and \textit{centroid location}~(see supplementary material for definitions). They all investigate the compactness of an object, hence, its spatial distribution. Compactness is highly relevant as it can measure how far an object's shape is from regular, blobby shapes like a box or a circle, which might be easy to segment. In contrast, highly irregular object shapes with fine structures can reduce segmentation quality, as some parts are easily missed.

The object properties in the second category are \textit{distance}, \textit{hull}, and \textit{solidity}~(see supplementary material for definitions), which all assess an object's convexity. Hence, they measure if an object features structures that substantially stand out from the main object body, which might negatively impact the segmentation performance. Additionally, the convexity of an object is also known to play a vital role in human object perception, with convex objects being easier to recognize~\cite{kanizsa1976convexity}.

The third category of object properties, \textit{good continuation}, \textit{symmetry}, and \textit{connectedness}~(see supplementary material for definitions), collects further shape properties. While symmetry or good continuation support object perception in humans according to Gestalt theory~\cite{wertheimer1923,kanizsa1976convexity}, the other property investigates whether an object is split into spatially distinct components, which makes recognizing the entire object more complex.

Finally, the three additional object properties -- \textit{centroid proximity}, \textit{color contrast}, and \textit{object color}~(see supplementary material for definitions) -- measure the interaction of objects or the object and the rest of the image.

\subsubsection{Study on Annotation Schemes}

To evaluate an object property, we calculate the property for all objects in COCO \textit{train} and split the annotations into three equally sized groups according to the calculated values. Note that connectedness is an exception to this, as it is only binary. This way, we generate a group of annotations with low values for a specific property, i.e., below a certain threshold, as well as group for annotations with medium, and high values, respectively. For instance, we assign objects with a center distance~(compactness) below 0.392 to the group of low compactness objects, while objects with a center distance above 0.598 are in the high compactness group. Subsequently, we assign each group to an annotation level and train AnnsMixer with these annotations. When assigning the groups to the annotation levels, we assign the group, which is presumably simpler to segment, to a weaker annotation level and vice versa. For instance, objects with a strong color contrast w.r.t. the surrounding do not profit as much from mask supervision compared to objects with low color contrast, where SAM has difficulties recovering the shape from a box or point. Hence, such groups of objects are assigned to point or box level. 

\begin{figure}
    \centering
    \begin{tabular}{ccc|ccc}
      \includegraphics[height=1.8cm]{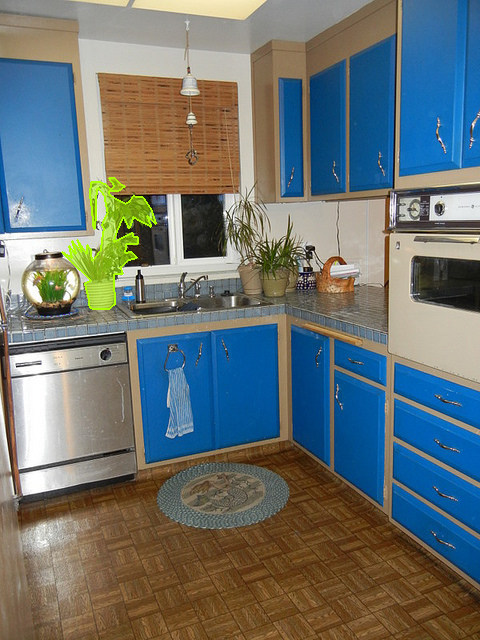} & \includegraphics[height=1.8cm]{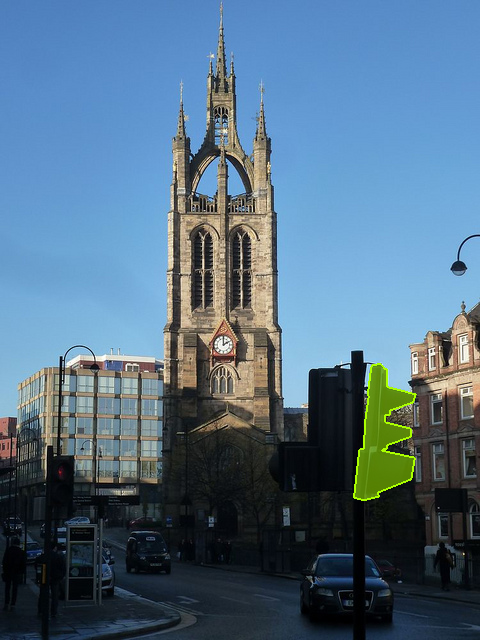} & \includegraphics[height=1.8cm]{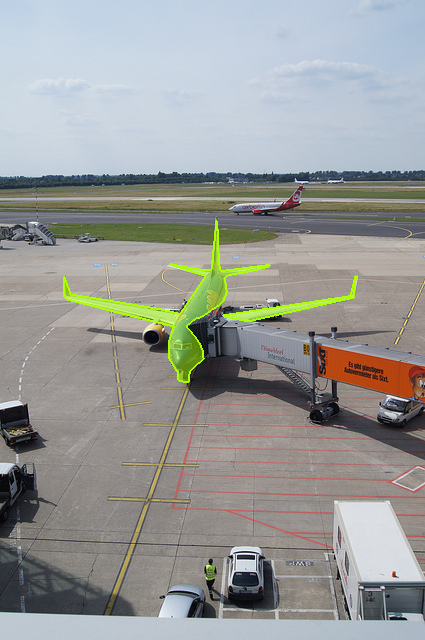} & \includegraphics[height=1.8cm]{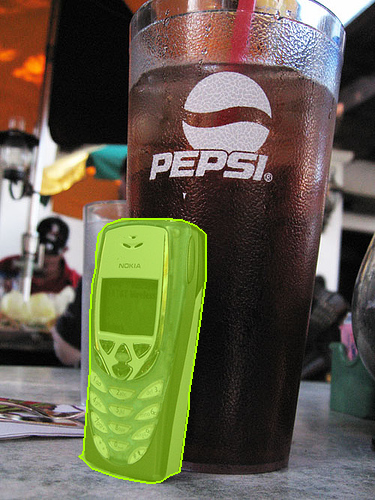} & \includegraphics[height=1.8cm]{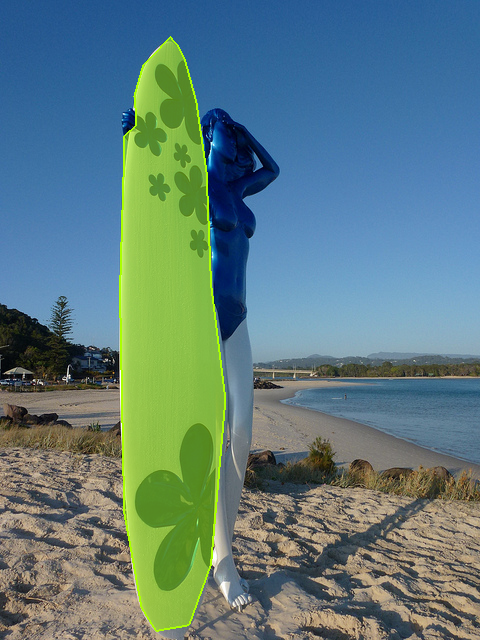} & \includegraphics[height=1.8cm]{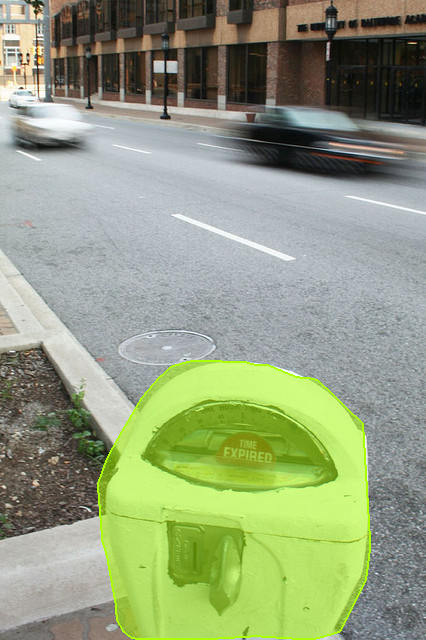} \\
    \includegraphics[height=1.2cm]{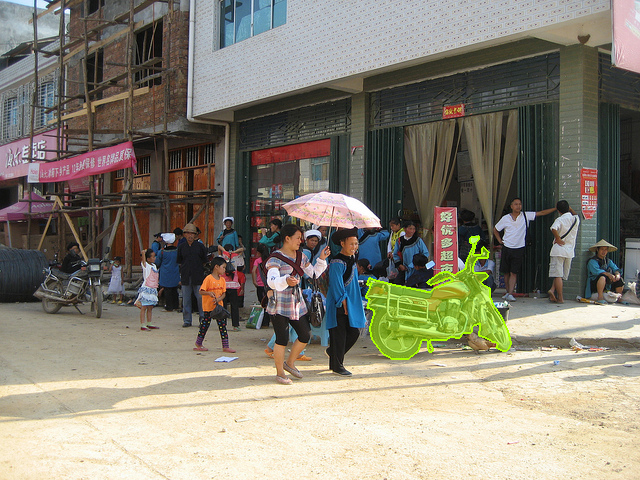} & \includegraphics[height=1.2cm]{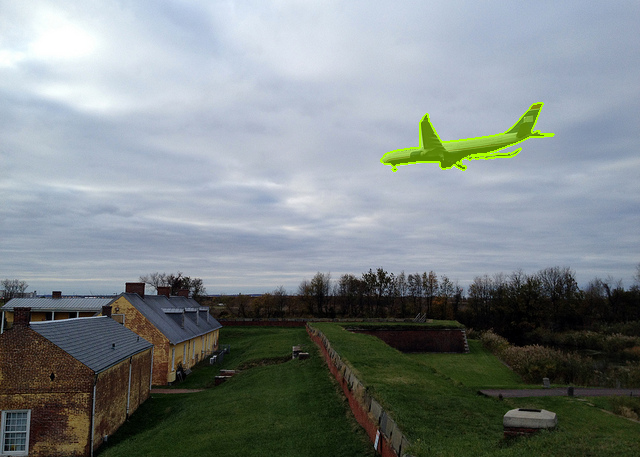} & \includegraphics[height=1.2cm]{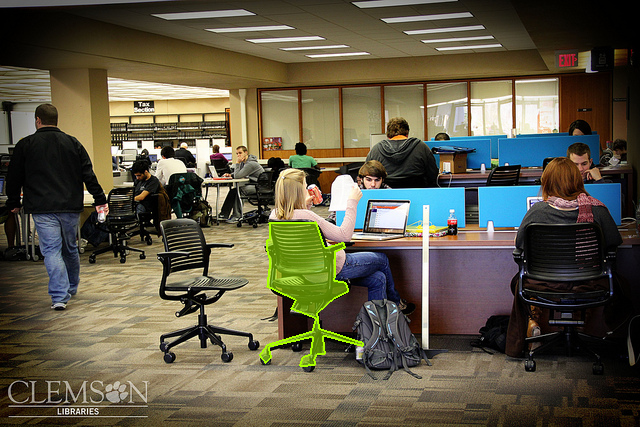} & \includegraphics[height=1.2cm]{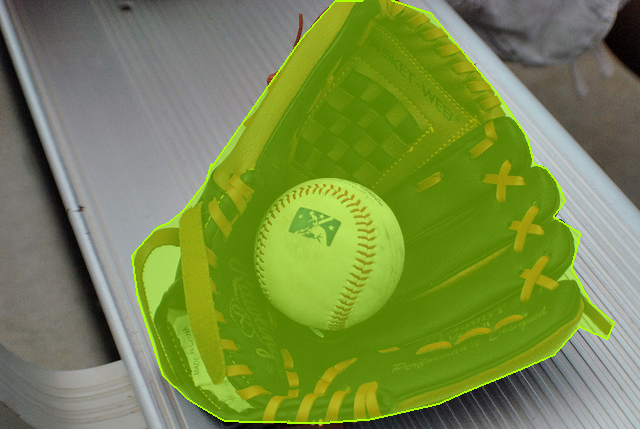} & \includegraphics[height=1.2cm]{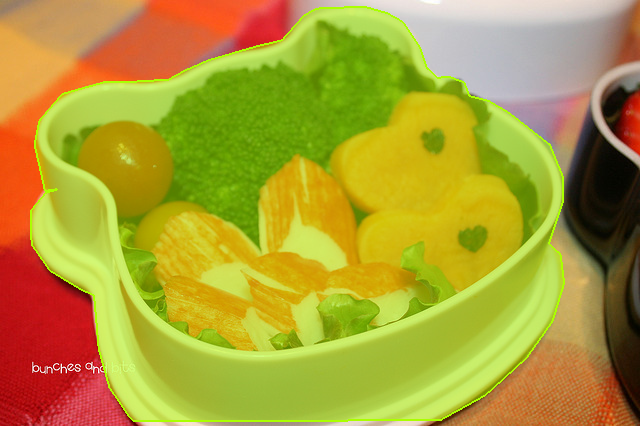} & \includegraphics[height=1.2cm]{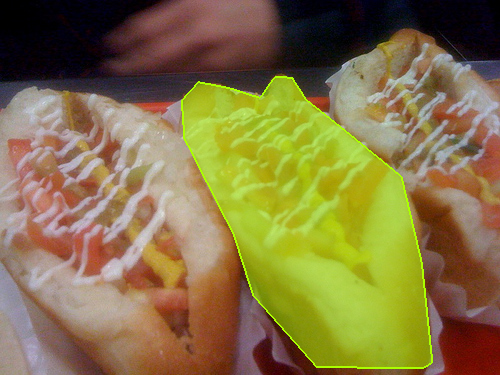} \\
    \multicolumn{3}{c}{Low good continuation} & \multicolumn{3}{c}{High good continuation}
    \end{tabular}
    \caption{Examples for objects with low good continuation~(left), which will be annotated with boxes in our Mix\textsubscript{efficient}, and objects with high good continuation~(right), which will be annotated with a point in our Mix\textsubscript{efficient}.}
    \label{fig:visGoodCont}
\end{figure}

Overall, we evaluated annotation schemes consisting of one or two object properties. Taking more object properties will presumably difficult to handle for human annotators. For each annotation scheme, we trained AnnsMixer and evaluated it on the COCO \textit{val} split. This study yielded the final annotation scheme Mix\textsubscript{efficient}. This mix utilizes only box and point annotations. Point annotations are used for objects with a smooth contour~(good continuation property), which fits to typical SAM segments based on points. For all other objects, boxes are used. This substantially reduces the annotation time, while keeping the required object property simple to spot. See also Fig.~\ref{fig:visGoodCont} for examples of objects with high and low good continuation. Given the annotation times in~\cite{kim2023devil}, annotating an image in the COCO dataset with the Mix\textsubscript{efficient} scheme only takes 114.7\,s on average compared to 650.8\,s in the fully-supervised case. Hence, our annotation scheme combines simplicity with strong annotation time reduction.

\subsubsection{Results}

To validate our efficient annotation scheme and test the generalization ability of Mix\textsubscript{efficient} beyond COCO, we compare fully-supervised Mask R-CNN and AnnsMixer trained with Mix\textsubscript{efficient} on four common instance segmentation datasets: COCO, PASCAL VOC~\cite{everingham2010pascal}, UVO~\cite{wang2021unidentified}, and Cityscapes~\cite{cordts2016cityscapes}. For each dataset, we train on the respective training split and validate on the datasets' validation or test splits. The results in Tab.~\ref{tab:annsMix} show that the annotation time\footnote{We adapt the calculations of~\cite{kim2023devil} to the respective datasets to determine the annotation time.} is reduced by up to 90\%~(UVO and Cityscapes), while the segmentation performance is only slightly degraded~(-5\% on average). For UVO, the mAP even improves when applying our annotation scheme. We suspect that this is the result of more precise annotations. The overall results also indicate a strong generalization of Mix\textsubscript{efficient} across the four diverse datasets.  To validate that Mix\textsubscript{efficient} is a well-designed annotation scheme, we compare it to a random mix with the same amount of point and box annotations~(Mix\textsubscript{random}). As visible from Tab.~\ref{tab:annsMix}, the mAP drops by up to 11\% from Mix\textsubscript{efficient} to Mix\textsubscript{random}, which highlights the strengths of our Mix\textsubscript{efficient}. Note that these annotation schemes are only possible using AnnsMixer's unique feature to handle diverse annotation levels within an image.  

\begin{table}[]
\centering
\caption{Results of Mask R-CNN and AnnsMixer trained with full supervision~($\mathcal{M}$), our proposed efficient annotation scheme Mix\textsubscript{efficient}, and the random annotation scheme Mix\textsubscript{random} on four different datasets. Annotation times are estimated per image based on~\cite{kim2023devil}. Colors of relative differences indicate improvement~(green) or decline~(red).}
\label{tab:annsMix}
{
    \begin{tabular}{@{}llccc@{}}
    \toprule
     & & Mask R-CNN & \begin{tabular}[c]{@{}l@{}}AnnsMixer with\\ Mix\textsubscript{efficient}\end{tabular} & \begin{tabular}[c]{@{}l@{}}AnnsMixer with\\ Mix\textsubscript{random}\end{tabular} \\ \cmidrule(l){3-5}
    Dataset & Metric & $\mathcal{M}$ & $\mathcal{B}+\mathcal{P}$ & $\mathcal{B}+\mathcal{P}$ \\ \midrule
    \multirow{2}{*}{COCO} & mAP & 0.366 & 0.347~(\textcolor{red}{-5.2\%}) & 0.325~(\textcolor{red}{-11.2\%}) \\
     & Ann. time & 650.8\,s & 114.7\,s~(\textcolor{mygreen}{-82.4\%}) & 114.7\,s~(\textcolor{mygreen}{-82.4\%}) \\ \midrule
    \multirow{2}{*}{VOC} & mAP & 0.302 & 0.275~(\textcolor{red}{-8.9\%}) & 0.267~(\textcolor{red}{11.6\%}) \\
     & Ann. time & 207.8\,s & 32.1\,s~(\textcolor{mygreen}{-84.6\%}) & 32.1\,s~(\textcolor{mygreen}{-84.6\%}) \\ \midrule
    \multirow{2}{*}{UVO} & mAP & 0.125 & 0.127~(\textcolor{mygreen}{+1.6\%}) & 0.113~(\textcolor{red}{-9.6\%}) \\
     & Ann. time & 780.8\,s & 57.2\,s~(\textcolor{mygreen}{-89.4\%}) & 57.2\,s~(\textcolor{mygreen}{-89.4\%}) \\ \midrule
    \multirow{2}{*}{Cityscapes} & mAP & 0.320 & 0.290~(\textcolor{red}{-9.4\%}) & 0.266~(\textcolor{red}{-16.9\%}) \\
     & Ann. time & 1440.4\,s & 101.7\,s~(\textcolor{mygreen}{-92.9\%}) & 101.7\,s~(\textcolor{mygreen}{-92.9\%}) \\ \bottomrule
    \end{tabular}
}
\end{table}

\begin{figure}
    \centering
    \includegraphics[width=0.6\linewidth]{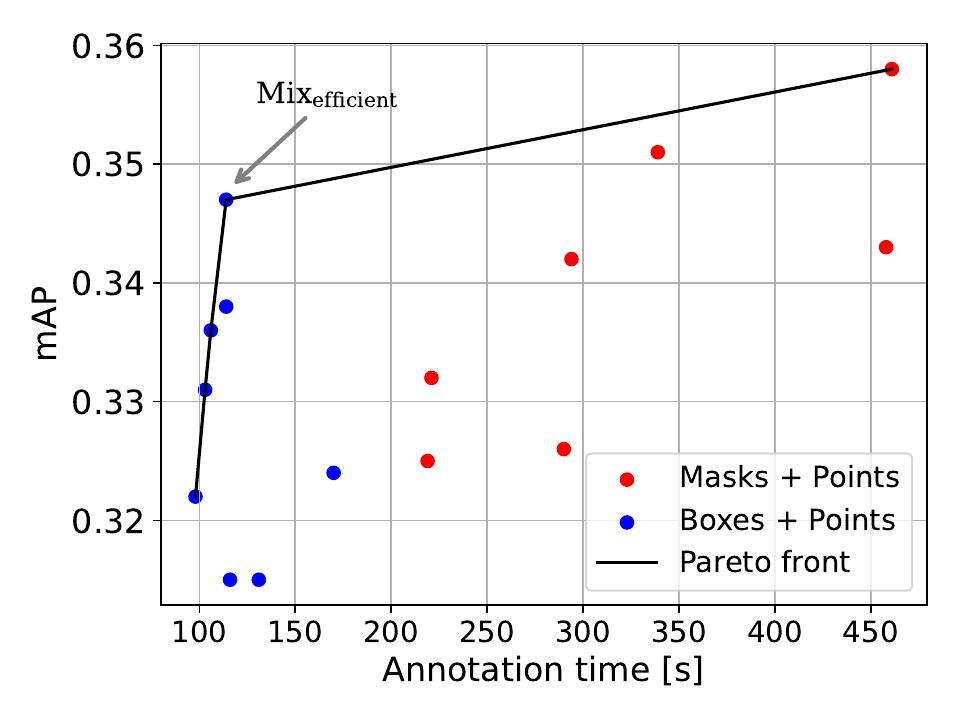}
        \caption{Results of AnnsMixer trained various annotation schemes evaluated during our study. Annotation times are estimated per image based on~\cite{kim2023devil}.}
    \label{fig:pareto}
\end{figure}

Finally, we show in Fig.~\ref{fig:pareto} an overview of AnnsMixer results on COCO \textit{val} based on different annotations schemes we constructed as part of this study. Since we are solving an optimization problem with two aims, strong segmentation performance in terms of mAP and low annotation time, a single best solution is difficult to produce. Therefore, Fig.~\ref{fig:pareto} shows the results of various annotation schemes with one or two rules based on the object properties discussed above. It is clearly visible that Mix\textsubscript{efficient} is the best trade-off, while other annotation schemes lead to slightly better segmentation performance or shorter annotation time. To better emphasize the multitude of strong annotation schemes for various scenarios, e.g., very low annotation time,  Fig.~\ref{fig:pareto} also shows the Pareto front.


\subsection{Annotation Merging with MixSIS}

As a second case study, we combine heterogenous annotation levels across different splits of a dataset. The goal is to substantially increase the amount of training data.  We test this idea on the datasets PASCAL VOC~\cite{everingham2010pascal} and BDD100k~\cite{yu2020bdd100k} with their instance segmentation and object detection annotations. PASCAL VOC's training dataset includes 6,929 mask and 27,450  box annotations, while BDD100k has 89,098 mask and 822,072 box annotations. For both datasets, we train on the respective training sets and evaluate on the validation sets. We follow our MixSIS formulation using mask and box annotations for AnnsMixer, while Mask R-CNN only utilizes the mask annotations.  

\begin{table}[]
\centering
\caption{Results of Mask R-CNN and AnnsMixer using mask and box annotations on the  PASCAL VOC dataset. Best results  are in bold.}
\label{tab:mix_pascal}
{
\begin{tabular}{@{}lcccc@{}}
\toprule
System     & Supervision         & mAP & mAP$_{50}$ & mAP$_{75}$ \\ \midrule
Mask R-CNN & $\mathcal{M}$       & 0.288 & 0.530 & 0.288 \\ 
AnnsMixer (our)     & $\mathcal{M}$+$\mathcal{B}$              & \textbf{0.385} & \textbf{0.660} & \textbf{0.401} \\ 
\bottomrule
\end{tabular}
}
\end{table}

Table~\ref{tab:mix_pascal} displays the results on the PASCAL VOC dataset. Given the additional box annotations, our AnnsMixer outperforms Mask R-CNN based on mask annotations by 0.097 in mAP~(+34\%). This indicates the strong positive influence of the additional images with box annotations. Note that also in the precise segmentation results~(mAP$_{75}$), AnnsMixer with additional box annotations outperforms Mask R-CNN relying only on mask annotations. Hence, the pseudo masks derived from the boxes in AnnsMixer have a high quality. A similar behavior is visible on the BDD100k dataset. The results in Tab.~\ref{tab:mix_bdd} show a 0.037 improvement in mAP of AnnsMixer using boxes and masks over vanilla Mask R-CNN. While the improvement is lower compared to PASCAL VOC, a strong positive influence of the additional box annotations is clearly visible.

\begin{table}[]
\centering
\caption{Results of Mask R-CNN and AnnsMixer using mask and box annotations on the BDD100k dataset. Best results  are in bold.}
\label{tab:mix_bdd}
{
\begin{tabular}{@{}lcccc@{}}
\toprule
System     & Supervision         & mAP & mAP$_{50}$ & mAP$_{75}$ \\ \midrule
Mask R-CNN & $\mathcal{M}$       & 0.229 & 0.423 & 0.216 \\
AnnsMixer (our)     & $\mathcal{M}$+$\mathcal{B}$              & \textbf{0.263} & \textbf{0.455} & \textbf{0.261} \\
\bottomrule
\end{tabular}
}
\end{table}

\subsection{Dataset Merging with MixSIS}

The final case study takes the annotation merging one step further. Here, we merge datasets on the same or overlapping recognition tasks, which feature different annotation types. The goal is, similar to the annotation merging, increasing the amount of training data for a specific task. Note that we do not consider any kind of domain adaptation and leave this for future work since we focus on a proof-of-concept. In this study, we select two recognition task: (a) apple detection and (b) general instance segmentation. 

For the apple detection, we merge the dataset MinneApple~\cite{hani2020minneapple}, which features 670 training images with 28,183 apples annotated with masks, and the MAD dataset~\cite{JohansonEtAlWACV2024} with 105 images containing 14,667 apples annotated with boxes. As a baseline, we train Mask R-CNN only on MinneApple's training set and evaluate on its test set. Leveraging the additional knowledge of the MAD dataset, we train our AnnsMixer with both mask annotations from MinneApple and box annotations from the entire MAD dataset. Subsequently, we also evaluate it on MinneApple's test set. The results in Tab.~\ref{tab:apple} show that despite the absence of domain adaptation and a substantial domain gap (industrial orchard vs. freestanding trees, apple density, and apple size), the additional training data from MAD in AnnsMixer already improves the mAP by about 2\%.

\begin{table}[]
\centering
\caption{Results of Mask R-CNN using MinneApple's mask annotations and AnnsMixer using MinneApple's mask and MAD's box annotations on the MinneApple test set. Best results are in bold.}
\label{tab:apple}
\begin{tabular}{@{}lcccc@{}}
\toprule
System     & Datasets         & mAP & mAP$_{50}$ & mAP$_{75}$ \\ \midrule
Mask R-CNN & MinneApple ($\mathcal{M}$)      & 0.360 & 0.722 & 0.322 \\ 
AnnsMixer (our)     & MinneApple ($\mathcal{M}$) + MAD ($\mathcal{B}$) & \textbf{0.367} & \textbf{0.723} & \textbf{0.330} \\ \bottomrule
\end{tabular}
\end{table}

For general instance segmentation, we merge the 6,929 mask annotations from PASCAL VOC's~\cite{everingham2010pascal} training dataset with the 492,812 annotations of the COCO training dataset, which overlap with the 20 object classes in PASCAL VOC. However, we do not take the mask annotations from COCO, but downgrade them to image-level annotations. Doing so allows us to study a practical case that utilizes image-level annotations in the MixSIS setup. The results in Tab.~\ref{tab:pascal_coco} clearly show the benefit of using abundant image-level annotations in this case. The results in mAP improve by $11\%$ from only using mask annotations to combining mask and image-level annotations in a MixSIS setting using our AnnsMixer. Note that while this particular case might be artificial, it shows that image-level annotations are also relevant in MixSIS and should not be overlooked.

\begin{table}[]
\centering
\caption{Results of Mask R-CNN using PASCAL VOC's mask annotations and AnnsMixer using PASCAL VOC's mask and COCO's image-level annotations on the PASCAL VOC validation set. Best results are in bold.}
\label{tab:pascal_coco}
\begin{tabular}{@{}lcccc@{}}
\toprule
System     & Datasets         & mAP & mAP$_{50}$ & mAP$_{75}$ \\ \midrule
Mask R-CNN & PASCAL VOC ($\mathcal{M}$)       & 0.294 & 0.513 & 0.306 \\ 
AnnsMixer (our)     & PASCAL VOC ($\mathcal{M}$) + COCO ($\mathcal{I}$) & \textbf{0.325} & \textbf{0.546} & \textbf{0.341} \\ \bottomrule
\end{tabular}
\end{table}

\section{Conclusion}

In this paper, we introduced a new task, MixSIS, to make the traditional WSIS formulation more flexible by allowing a mixture of different annotation levels even within a single image. To address MixSIS, we proposed the pioneering baseline system AnnsMixer, which produces pseudo annotation masks for all annotation levels in a single system. Our evaluation demonstrates that AnnsMixer largely outperforms state-of-the-art systems in WSIS and MixSIS. To highlight the practical relevance of the MixSIS task, we presented three case studies, leading to reduced annotation time~(-90\%) or improved the segmentation quality (+34\%) through leveraging AnnsMixer's unique features. Overall, MixSIS has a strong potential to aid practical instance segmentation applications by reducing the annotation effort.

\bibliography{main}

\end{document}